\documentclass{spie}

\usepackage{cite}

\usepackage{times} 
\usepackage{indentfirst} 

\usepackage[colorinlistoftodos,color=pink]{todonotes} 
\usepackage{xkeyval}
\presetkeys{todonotes}{inline}{}

\usepackage{booktabs} 
\usepackage{floatrow}

\usepackage{lipsum} 

\title{ SDE-AWB: a Generic Solution for 2nd International Illumination Estimation Challenge 
}

\author{Yanlin Qian\supit{1,2}, Sibo Feng\supit{1}, Kang Qian\supit{1}, 		
        Miaofeng Wang\supit{1}
  \skiplinehalf
  \normalsize 
  \supit{1}Huawei Technology \\
  \supit{2}Tampere University \\
}

\begin{document}

\maketitle

\begin{abstract}
  
  We propose a neural network-based solution for three different tracks of 2nd International Illumination Estimation Challenge (chromaticity.iitp.ru). Our method is built on pre-trained Squeeze-Net backbone, differential 2D chroma histogram layer and a shallow MLP utilizing Exif information. By combining semantic feature, color feature and Exif metadata, the resulting method -- SDE-AWB -- obtains 1st place in both indoor and two-illuminant tracks and 2nd place in general track. 
  
\end{abstract}

\section{Introduction}

Auto-white balance, in short AWB, is a necessary processing step in the image signal processor (ISP) in most digital cameras.  This function originates from the human eye's property -- color constancy providing object coloration recognition independent of the casting scene illumination. This problem has been heavily explored over decades \cite{gijsenij2011computational}.

Starting from the classic Gray World algorithm \cite{buchsbaum1980spatial} assuming the global averaged vector of the captured scene is achromatic, a large selection of AWB methods have been proposed, including but not limited to Gray Edge \cite{van2007edge}, Gray Pixels \cite{yanlin2019vissap, yanlin2019arxiv}, Random Forest-based AWB \cite{cheng2015effective}, Corrected-Moment \cite{finlayson2013corrected} and Convolutional Color Constancy \cite{barron2015convolutional, barron2017fourier}. In the era of deep convolutional neural network (CNN), there are also a list of works that challenges the conventional works and refreshes the state-of-the-art performance from year to year. Bianco \textit{et al.} \cite{bianco2015color} initially shows the potential of applying CNN for illumination estimation. Then more effective and efficient CNN variants are designed, like AlexNet+SVM \cite{yanlin2016icpr}, DS-Net \cite{shi2016eccv}, FC4-Net \cite{hu2017cvpr}, RCC \cite{yanlin2017iccv}, BCC \cite{qian2020benchmark} and so on. 

Along with the rapid changes in AWB work, the mainstream datasets for assessing AWB methods are varying as well. SFU laboratory dataset \cite{barnard2002data} and SFU gray ball dataset \cite{ciurea2003large} are two datasets published over 20 years. Both have their design merits, but are less visited nowadays due to the limited number of images or poor image resolution. In recent years, Gehler \textit{et al.} \cite{gehler2008bayesian, shi2010re} released Gehler-Shi dataset and Cheng \textit{et al.} \cite{Cheng14} published the NUS 8-camera dataset. Compared to the mentioned SFU data, both datasets contains more varying-scene images (568 for Gehler-Shi, ) are captured using DSLR cameras (its high resolution also allows cropping to generate more samples). The up-to-date works C4-Net \cite{yu2020aaai}, ICDF \cite{xu2020end} achieved nearly-saturated results over these two datasets. There exists a great need of a new challenging large-scale dataset. 
The 2nd illumination estimation challenge, which this paper is submitted for, brings such a dataset for AWB study. For convenience, we call this challenge dataset as IITP dataset \footnote{credits to IITP (iitp.ru) for the collection and label annotation.}.  The main attributes of IITP datasets are: in total it has nearly five thousand DSLR images, each of which annotated with two illuminant vectors and Exif tag; All images are divided into three tracks (general, indoor and two-illuminant cases), with regarding to the captured scenes and the discrepancy of the measured two illuminant vectors. 

For the challenge, we propose a net, called SDE-AWB, for all three tracks. To allow the method to output single illuminant vector or two, SDE-AWB itself is a flexible combination of three different learnable modules:  Squeeze-Net backbone, Differential 2D chroma histogram layer and a shallow Exif MLP.

\section{methodology}

In this section we describe the overall design of our proposed SDE-AWB, in the form of its components. Each component can be treated as a single neural network, which can do a complete illumination estimation job with some necessary regression layers. By varying the combination of different components, we get different model variants for single-illuminant and two-illuminant tracks.

\begin{figure}[h]
    \centering
    \includegraphics[width=0.70\linewidth]{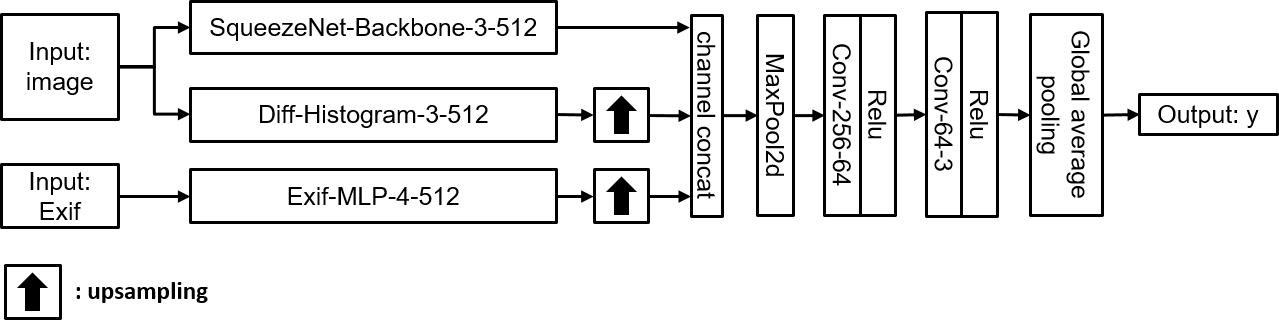}
    \caption{ The architecture of SDE-AWB (model A). 
``LayerName-$x$-$y$'' denotes a 2D layer with $x$ input channels and $y$ output channels where the layer
is either a standard convolution layer,
a backbone network (e.g. SqueezeNet) or
a 2D LSTM. ``umsampling'' does the upsampling to match the spatial dimensions of the output of SqueezeNet backbone. ``channel concat'' concatenates three features maps along the feature channel axis. 
$\mathbf{y}$ is the illumination color
vector after the last ReLU layer.
    }
    \label{fig:architectureA}
    \vspace{-2mm}
\end{figure}

\noindent \textbf{Model A:} The basic architecture (model A) of SDE-AWB is depicted in Figure~\ref{fig:architectureA}. Model A has 3 sub-branches for extracting different types of representation from pictorial domain or ancillary metadata domain (Exif info).  The final layer of model A applies global average pooling over 3-channel feature maps, outputting a global illumination vector, for the single-illuminant track.

\begin{figure}[h]
    \centering
    \includegraphics[width=0.70\linewidth]{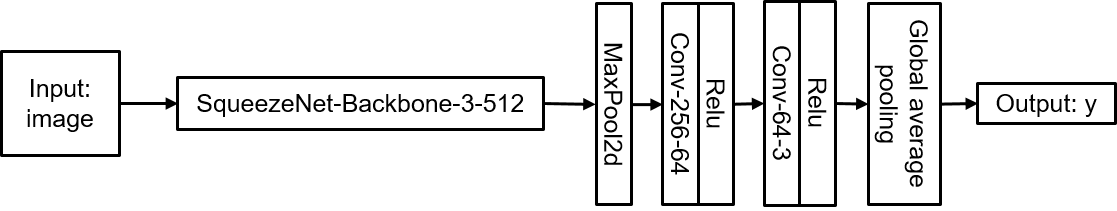}
    \caption{ The architecture of SDE-AWB (model B). 
    }
    \label{fig:architectureB}
    \vspace{-2mm}
\end{figure}

\noindent \textbf{Model B:} Simply deleting the sub-branches of Diff-Histogram and Exif-MLP (and concat layer), Model A turns into Model B shown in Figure~\ref{fig:architectureB}, which is actually the architecture of FC4-Net \cite{hu2017cvpr}. 

\begin{figure}[h]
    \centering
    \includegraphics[width=0.70\linewidth]{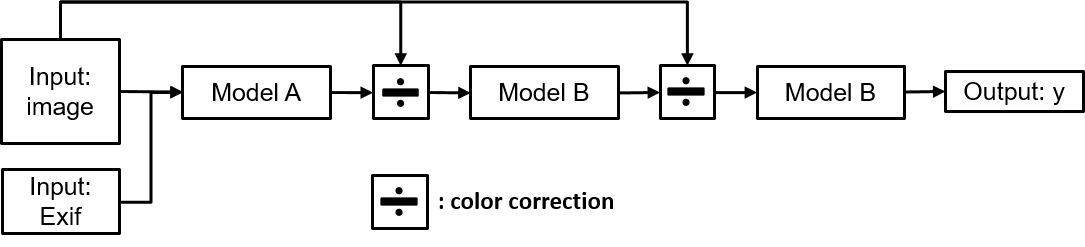}
    \caption{ The architecture of SDE-AWB (model C). ``color correction'' takes the predicted illumination vector from the layer above and do color correction on the input image, resulting into a color-corrected image. Note that model A takes 2 inputs while model B takes only 1 input.  
    }
    \label{fig:architectureC}
    \vspace{-2mm}
\end{figure}

\noindent \textbf{Model C:} With the predicted illumination vector of Model A, we can multiply the input image with a color correction matrix to do ``white-balance'', giving another image. In such way, as shown in Figure~\ref{fig:architectureC}, we concatenate Model A with 2 Model Bs. This cascaded network, we call Model C, is used for our submissions in general track and indoor track. 

Replacing the last Conv-64-3 in Model C by Conv-64-6, the model can generate two illuminant vectors and is used for two-illuminant track. 
In the following, we mainly detail three key components.  


\subsection{Squeeze-Net Backbone $\times 3$}
SqueezeNet was initially proposed in \cite{iandola2016squeezenet} for light-weight image classification. SqueezeNet relys on its novel ``fire module'' to massively decrease the number of parameters without hindering the performance.  For the detail of SqueezeNet, we refer readers to the original paper.   

Hu \textit{et al.} \cite{hu2017cvpr} introduced SqueezeNet to the task of AWB and achieved top-performing results over Gehler-Shi dataset and NUS 8-camera dataset. In \cite{hu2017cvpr}, the layers from ``conv1'' to ``fire8'' (plus an extra $2\times2$ pooling) in SqueezeNet are preserved to realize fully convolutional inference. To mapping high-dimensional features to the global illumination vector, two more conv layers and global averaging pooling are applied.  Then Yu \textit{et al.} \cite{yu2020aaai} cascades up to three SqueezeNets to realize a coarse-to-fine illumination regression, which inspires us when designing Model C on basis of Model A.

\subsection{Differential Histogram}
Different to hard-coded histogram operation like that implemented in opencv, Hu \textit{et al.} \cite{xu2020end} introduces differential color histogram to the task of AWB. Differential Histogram has two benefits: 1) by training with other cascaded neural network layers, it allows histogram bins to be tuned automatically in terms of location and size; 2) fast inference on GPU grid.  Making a histogram over an image is a voting procedure, formulated as: 

\begin{equation}
    \Psi_{k,b}(x_k) = max(0, 1-|x_k - \mu_{b}| \times \omega_{b}), 
  \label{eq:diffhist}
\end{equation}
where $\Psi_{k,b}(x_k)$ refers to the value of $k$-th element in in feature map $x$, for the $b$-th histogram bin. $\mu_{k,b}$ and  $\omega_{k,b}$ are the center value and the width value of the $b$-th histogram bin, respectively.  Giving a feature map $x$ with shape of $C \times H\times W  \times 1$, the output of Diff-Histogram has the shape of $C \times H\times W  \times B$ \footnote{C:channel dimension, 3 for rgb image, H:height, W:width, B:the number of histogram bins}.  

Following \cite{xu2020end}, we utilize 4-scale spatial pyramid pooling to extract structural information from histogram feature. The pooling strides are set to be 1,2,4 and 8, respectively. All pooled histograms are flatten for depth-wise concatenation, followed by a Conv-$N_{hist}$-512 and ReLU.  $N_{hist}$ is the number of channel of the concatenated histogram features. 

\subsection{Exif Multilayer Perceptron}

Exif information records the image metadata when capturing an image, and is usually stored in raw data, for example, CR2 images for Canon cameras. Luckily, for each image in any tracks in the challenge, the exif is parsed and released. We propose a Exif MLP to learn a joint metadata feature from a list of exif items (in our model, we use aperture, exposure time, iso and orientation). The Exif-MLP is unfolded as [Conv-4-4, ReLU, Conv-4-512, ReLU].  As shown in Figure~\ref{fig:architecture}, by upsampling and channel concatenation, the Exif-MLP output can be merged int the main branch, affecting the inference of the final estimate in a postive way. 

\section{Experiments}

\subsection{Dataset Preprecessing}
Thanks to the challenge organizers for releasing the largest AWB datasets. About 5k images are collected using the same camera model (Canon 600D and Canon 550D), and for each of them, a double-face color cube is used to measure two light source vectors. If the angular error between the two measured vectors is not less than 2 degree, according to the challenge standard, the image is classified as a two-illuminant case.

\begin{figure}
    \centering
    \includegraphics[width=0.70\linewidth]{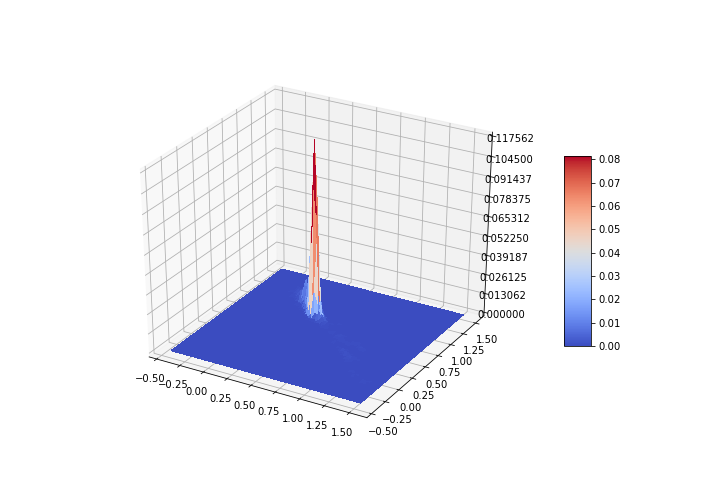}
    \caption{ Gaussian-blured UV histogram of General Trainset. 
    }
    \label{fig:uvhist_generaltrainset}
    \vspace{-2mm}
\end{figure}

Considering the Cube+ dataset is also collected using Canon 550D, we adopt it here to expand the datasets used for different tracks. If the angular error of two measured illuminant vectors is over 2 degrees, we assign the Cube+ image to the trainset of two-illuminant track. For the single-illuminant track, take general track as an example, we apply rgb2uv operation (Equation~\ref{eq:rgb2uv}) and make a 2D uv histogram, followed by a gaussian blur (shown in Figure~\ref{fig:uvhist_generaltrainset}).  The filter size of the gaussian kernel is set to be 7 in our method. 
Then, for each image in Cube+ dataset, if the uv of the groundtruth vector of this image falls into the scope of non-zero area of the precomputed uv histogram, we assign this image to the set ``Cube+ General'', which will be used for training our general-track model.  In such a way, we pick up a list of images from Cube+ dataset which fit different challenge tracks, without introducing more diversity in illumination labels. Table~\ref{tab:dataset} gives the trainset statistics for 3 tracks.

\begin{equation}
  u = log(g/r), v= log(g/b)
  \label{eq:rgb2uv}
\end{equation}

\begin{table}
\begin{center}
 \begin{tabular}{|c c c c|} 
 \hline
 Track & General & Indoor & Two-illuminant \\ [0.5ex] 
 \hline\hline
 Trainset1 & General Trainset (2661) & Indoor Trainset (329) & Two-illu Trainset (604) \\ 
 \hline
 Trainset2 & Cube+ General (1663) & Cube+ Indoor (1663) & Cube+ Two-illu (403) \\
 \hline
 Learning Rate & 3e-4 & 3e-4 & 3e-4 \\
 \hline
 Batch Size & 16 & 16 &16 \\
  \hline
 Epochs & 450 & 500 & 350 \\
 \hline
\end{tabular}
\end{center} 
\caption{Datasets and Training details for training three different models. The number in brackets refers to the number of images in the image set. }
\label{tab:dataset}
\end{table}

\subsection{Training Details}

Although in the challenge, different metrics will be used for evaluating the submitted results for 3 tracks. For example, for general track, worst-$25\%$ reproduction error is used while for two-illuminant track,
the metric is the minimal squared sum of two angular reproduction errors.
For simplicity and consistence, we adopt the same recovery angular error
for training our models. The metric we used is formulated as:

\begin{equation}
  \varepsilon_{\hat{{c}},{c}_{gt}} = \arccos\left(\frac{\hat{{c}}\cdot {c}_{gt}}{\parallel \hat{{c}}\parallel  \parallel {c}_{gt}\parallel}\right)  , 
    \varepsilon_{\hat{{c}},{c}_{gt}} = 0.5 \arccos\left(\frac{\hat{{c}_l}\cdot {c}_{gt,l}}{\parallel \hat{{c}_l}\parallel  \parallel {c}_{gt,l}\parallel}\right) + 0.5 \arccos\left(\frac{\hat{{c_r}}\cdot {c}_{gt,r}}{\parallel \hat{{c_r}}\parallel  \parallel {c}_{gt,r}\parallel}\right)  , 
  \label{eq:metric}
\end{equation}

where $\cdot$ denotes the inner product between the two vectors and $\parallel\parallel$ is the Euclidean norm. $\hat{{c}}$ is the predicted illumination vector and  ${c}_{gt}$ the ground truth vector. For two-illuminant case, the predicted vectors are the left vector $\hat{{c_l}}$ and the right vector $\hat{{c_r}}$.

In data augmentation, $512\times 512$ patches are cropped from the orignal raw image.  The patch is then augmented using a random rotation with angle in [-60$^\circ$, 60$^\circ$] and resize ratio in [0.1, 1.0]. We use Adam optimizer to all models. The training scheme is detailed in Table~\ref{tab:dataset}. 



\subsection{Results by Challenge Online Server}

\begin{table}
\scriptsize
\begin{center}
\caption{General Track Leaderboard. The notation $\Uparrow(n)$ refers to the method being ranked $n$th. The numbers on IITP dataset are retrieved from the challenge page as the test data is agnostic to the challenge players. } 
\label{tab:general}
  \begin{tabular}{l c c c c}
    \toprule 
  &  \multicolumn{3}{c}{Undisclosed Testing Data }  \\
 Method & Worst-25\% & Mean & Median & Trimean  \\
 \midrule 
$\Uparrow(1)~$ CAUnet & 4.084 & 1.605 & 0.966 & 1.084    \\
$\Uparrow(2)~$ SDE-AWB (model C) & 4.979 & 1.914 & 1.164 & 1.269 \\
$\Uparrow(3)~$ illumGAN & 9.999 & 4.643 &3.588 & 3.841 \\
$\Uparrow(4)~$ GreyWorld  & 10.419 & 4.500 & 3.319 & 3.611\\
$\Uparrow(5)~$ Const   & 17.023 & 7.081 &4.020 & 5.275\\
\bottomrule
  \end{tabular}
\end{center}\vspace{-3mm}
\end{table}

\begin{table}
\scriptsize
\begin{center}
\caption{Indoor Track Leaderboard. } 
\label{tab:indoor}
  \begin{tabular}{l c c c}
    \toprule 
  &  \multicolumn{3}{c}{Undisclosed Testing Data }  \\
 Method & Mean & Median & Trimean  \\
 \midrule 
$\Uparrow(1)~$ SDE-AWB (model C)  & 2.541 & 1.763 & 1.943  \\
$\Uparrow(2)~$ illumGAN & 3.191 & 2.596 & 2.674  \\
$\Uparrow(3)~$ PCGAN,MCGAN & 3.301 & 2.312 & 2.298  \\
$\Uparrow(4)~$ GreyWorld  & 4.106 & 3.673 & 3.545\\
$\Uparrow(5)~$ Const   & 15.270 & 14.802 & 15.332\\
\bottomrule
  \end{tabular}
\end{center}\vspace{-3mm}
\end{table}

\begin{table}[h!]
\scriptsize
\begin{center}
\caption{Two-illuminant Track Leaderboard. } 
\label{tab:twoillu}
  \begin{tabular}{l c c c c}
    \toprule 
  &  \multicolumn{3}{c}{Undisclosed Testing Data }  \\
 Method & Mean squared & Mean & Median & Trimean  \\
 \midrule 
$\Uparrow(1)~$ SDE-AWB (model A) & 31.026 & 2.751 & 2.262 & 2.290 \\
$\Uparrow(2)~$ 3du-awb & 37.305 & 2.863 & 2.503 & 2.497 \\
$\Uparrow(3)~$ GreyWorld  & 81.841 & 4.127 & 3.538 & 3.715\\
$\Uparrow(4)~$ Const   & 144.745 & 5.264 &3.475 & 3.815\\
\bottomrule
  \end{tabular}
\end{center}\vspace{-3mm}
\end{table}

Table~\ref{tab:general} gives the leaderboard for the general track, made by a online evaluation server. Our proposed method obtains the second place, which is closer to the top submission -- CAUnet compared to other methods . 
Table~\ref{tab:indoor} is the indoor-track leaderboard, where SDE-AWB ranks first and leads by a wide margin. Compared to the general track, the indoor track has no daylight illuminant (thus the blackbody curve cannot be relied on), making it more difficult to solve. The superiority of SDE-AWB mainly comes from three SqueezeNet backbones, diff-histogram and Exif Module, which will be proved by an ablation study.  Larger training set we use for indoor track is also a plus. 
Table~\ref{tab:twoillu} shows the results of the two-illuminant track. Our main competitor here is 3du-awb, which is surpassed by SDE-AWB in all reported error metrics. 

\subsection{Ablation Study}

\begin{table}[h!]
\scriptsize
\begin{center}
\caption{Ablation study on the introduced modules. The values in table are mean recovery angular errors in different settings.  } 
\label{tab:abaltion}
  \begin{tabular}{l c c c}
    \toprule 
  &  \multicolumn{3}{c}{Undisclosed Testing Data }  \\
 Method & General & Indoor & Two-illuminant  \\
 \midrule 
model B (SqueezeNet $\times 1$)  & 1.00 & 1.25 & 1.76  \\
model B (SqueezeNet $\times 3$)  & 0.48 & 1.21 & 1.35  \\
model A (SqueezeNet $\times 1$, Diff-Hist)  & 0.83 & -- & --  \\
model A (SqueezeNet $\times 1$, Diff-Hist, Exif)  & 0.79 & 1.06 & \textbf{1.19}  \\
model C (SqueezeNet $\times 3$, Diff-Hist, Exif)  & \textbf{0.43} & \textbf{0.99} & 1.37  \\
\bottomrule
  \end{tabular}
\end{center}\vspace{-3mm}
\end{table}

We evaluate the importance of each module of SDE-AWB. Considering the testset of the challenge is undisclosed, fro each track trainset, we randomly select 2/3 images for training and the remaining for testing. The profiled results are given in Table~\ref{tab:abaltion}. It is obviously model C achieves best performance over single-illuminant tracks, while for two-illuminant track, model A performs better than model C. We infer that the color correction step in cascaded network does not benefit learning location-sensitive illumination feature.  Therefore, we use model A for our final two-illuminant submission.  

\section{Conclusion}


In this paper we demonstrate a generic illumination estimation -- SDE-AWB, can well perform for general, indoor and two-illuminant tracks. With specific module design and combination, it obtains 1st place in both indoor (mean error 1.763) and two-illuminant (mean error 2.751) tracks and 2nd place in general track (mean error 1.914). Our used modules, Squeeze-Net backbone, differential 2D chroma histogram layer and a shallow MLP show their effectiveness on a cross-validated ablation study. Our future plan is to explore more elegant architecture to merge cross-domain information like Exif metadata.




\bibliographystyle{spiebib}
\bibliography{bibliography}





\end{document}